\documentclass[11pt,a4paper]{article}
\usepackage[hyperref]{acl2019}
\usepackage{times}
\usepackage{latexsym}
\usepackage{url}
\usepackage{graphicx}

\usepackage{multirow}
\usepackage{graphicx}
\usepackage[normalem]{ulem}
\useunder{\uline}{\ul}{}

\aclfinalcopy 

\title{Beyond Turing: Intelligent Agents Centered on the User}

\author{Maxine Eskenazi, Shikib Mehri, Evgeniia Razumovskaia and Tiancheng Zhao \\
  Dialog Research Center, Language Technologies Institute \\
  Carnegie Mellon University, USA \\
  \texttt{\{max,amehri,erazumov,tianchez\}@cs.cmu.edu}}

\date{}

\begin{document}
\maketitle
\begin{abstract}
Most research on intelligent agents centers on the agent and not on the user. We look at the origins of agent-centric research for slot-filling, gaming and chatbot agents. We then argue that it is important to concentrate more on the user. After reviewing relevant literature, some approaches for creating and assessing user-centric systems are proposed.
\end{abstract}

\section{Introduction}
Intelligent agents have entered the realm of popular use. Over 50,000,000 Alexa agents are in use today \citep{sterling2018}. As agents of all kinds become part of the household, the AI community must respect the individuals who use them. With close public scrutiny, the possibility of failure is high. If we are not attentive, our agents may cause us to lose public confidence. This can have serious consequences for research, such as drastically reduced funding, as we have seen in the case of automatic speech recognition (ASR) in the 1990s. Intelligent agents must live up to the users' expectations. Consequently, the research community should not set those expectations above what we can reasonably deliver. We should be especially concerned about the tendency of the community to concentrate on the agent and disregard the effect of the dialog on the end user. The two interlocutors are equals in this endeavor. The agent should be serving the myriad needs of the user in the same way that a car serves its user for travel, a pen for writing.

Agents presently serve users in three broad a-reas: accomplishing some task (slot-filling), providing enjoyment via gaming, and providing companionship (chat). Most research approaches these areas separately although some recent hybrid work such as DialPort~\cite{lee2017dialport} has combined slot-filling and chat. This paper attempts to address agents in all three areas. We also consider systems independently of whether they are embodied (since there are successful agents that are both embodied~\cite{stratou2017multisense,muessig2018didn,lycan2019direct} and not (Siri (Apple), Alexa (Amazon), etc)).

In the early days of dialog research, the user was treated on an equal par with the agent. The PARADISE~\cite{Walker1997PARADISEAF} approach was often used to assess system success. Researchers measured the length of a dialog and its success rate. Thus getting the user the information that had been requested was a sign of success. User feedback was also taken into account. 

Since the advent of PARADISE, the agent has been endowed with new qualities, for example, emotion~\cite{gratch2001tears}. The system chooses the ``correct'' emotion to express. It is assumed that if the system chooses the correct emotion, the user will benefit in some ill-defined way. Rapport and similar affective qualities~\cite{gratch2007creating} have also been added to some agents. Here again the goal is for the agent to correctly express a given emotion. If that happens, it is assumed that the result will be useful to the user. In both cases, there is little empirical proof that the new qualities have benefited the user. A few exceptions do exist. The reader is referred, for example, to the systems produced at USC-ICT~\footnote{\url{http://nld.ict.usc.edu/group/projects}}.

One of the overarching goals of the dialog/intelligent agent community has been to create an agent that is confused with a human, as defined by the Turing test. This idea originally appeared in Alan Turing's 1950 paper~\cite{turing1950computing}, although that was not the main concern of Turing’'s original premise. Computers had just begun to appear. Turing, like many at the time, was looking to answer the burning question of whether a computer would some day be able to ``think''. For this, he proposed that human thought had two characteristics: generalization and inference. Rather than testing each of the two separately, he sought an assessment scheme that accounted for both at the same time. While we could imagine several possible test scenarios, the one that Turing chose had two interlocutors hidden from a human judge; one being human and the other a computer. The human judge had to determine which one was the computer. To do this, the judge could ask each interlocutor a series of questions. The Turing test caught on thanks to the very seductive idea of a machine deceiving a human. Much research has gone into agents capable of making the user believe that they are human. Some of this research has produced agents that reside in the Uncanny Valley~\cite{moore2012bayesian}, having human qualities that cause discomfort to the user. Deception has created a sense of fear in the general public. Consequently much ``explaining'' has taken place (``robots are not out to take over the world, they will not replace humans''~\cite{lee2019}). The present paper will examine how we can get our research goals back on track: focusing on the user rather than on the appearance and functions of the agent. We believe that this change in focus will shape the design of new and powerful algorithms and agents in the future.

This paper will discuss existing work and then propose novel ways to approach the agent and its assessment. We address assessment since the manner in which we assess our agents has a strong influence on their development. The paper has six sections. Section 2 proposes a definition of the relation of the agent to the user in a user-centric approach. Section 3 gives background. Section 4 discusses the elephant in the room, the issue of getting users for training and testing. Section 5 proposes several novel approaches to assessing agents in a user-centric context and Section 6 concludes.

\section{Defining the role of the agent in a user-centric approach}
In order to focus on the user, we must first define the role of the agent. In 1950 Turing asked  ``Can the computer think?'' In the early 21st century, a more pertinent question is ``Can the computer serve?'' This question casts the agent as a \textit{partner} rather than an \textit{opponent}. The agent works seamlessly, hand in hand with the user, to arrive at some goal. All three types of agents (slot-filling, chat and hybrid) can be partners. The iPhone has become a partner. Apple stores often ask customers to come back after an hour to pick up a phone left for a simple repair. Employees observe that most customers come back in 15 minutes~\cite{apple} and are surprised to find that they have not been gone for a full hour. They appear to have lost the ability to estimate time without their iPhone. To get a more concrete definition of the agent as a partner, consider the well-known agent utterance, ``How may I help you?'' If the agent can begin a dialog asking for this guidance, then it is logical that the agent should end that dialog by asking, ``Have I helped you?''. The answer to this question is a first user-oriented assessment. This question is flexible enough to cover systems in all three areas. 
\begin{itemize}
    \item Have I helped you accomplish some task? (slot-filling)
    \item Has your interaction with me given you some enjoyment? (via gaming)
    \item Has your interaction with me given you some companionship? (via conversation - chat)
\end{itemize}

We can measure how well an agent fulfills this role from our personal observation, from third party observation (crowdsourcing, for example) and, especially, from the user’'s point of view (feedback via questionnaires, for example).  

Following are two examples of how a slot-filling agent can serve the user, going beyond providing a ``correct'' answer as output.

\subsection{Seniors}
\label{subsec:seniors}
Human capabilities change as we age. An agent partnering with a senior must be sensitive to slower information processing and less multitasking~\cite{black2002elderly}. Thus, for example, the agent must adapt its speaking rate to that of the senior user. It should also not provide seniors with more information than they can process at a given time.

An agent communicating with seniors should be assessed according to its capability to demonstrate these qualities. It should also be assessed according to what the user gets out of the interaction. Was the senior able to use the information in some way. Did they enjoy their chat with the agent? Would they use it again?

\subsection{Workers in industry}
Industry workers use agents for a variety of narrowly--defined applications. For example people inspecting equipment in factories use agents to record their observations. Such an agent must adapt to the user’'s background and preferences. Some users have had the same job for decades and need little help. The agent in this case mainly provides hands-free data entry. The advantage over a screen is that it enables the inspector to keep their eyes on the piece of equipment as they enter parameters into the system. In this case the person needs only one dialog turn to get something done: ``what is the measure on the upper right hand pressure gauge?'' A newer employee, on the other hand, needs more help, implicit training, until they are familiar with the task: ``now you need to measure the pressure in the upper right hand pressure gauge. That's the one in the green frame.'' ``Ok, I see it.'' ``You should hold down the button located under that gauge for 5 seconds.'' ``Ok, done.'' ``It will give you the correct reading as soon as you lift your finger.'' Some workers who have been on the job for a while may still need some help. The agent should be able to detect that. It should also know when the user needs extra time between turns. To assess an agent in this setting we determine whether the user was able to correctly finish their task, if they did it with less error, if they feel free to ask for extra information when needed and if they would use the agent again.

\section{Background}
In this Section, we discuss some of the literature on intelligent agents that do focus on the user. Significant effort has gone into the assessment of individual dialog system modules such as natural language generation (NLG). Liu et al~\shortcite{liu2016not} review dialog system assessment approaches. They describe word overlap and embedding-based strategies. They also provide evidence that existing measures are insufficient. The paper states that although embedding-based measures hold the most promise, no one has, to date, found a way to carry out unsupervised evaluation. In addition, while it is important to insure that modules function properly, authors cannot claim that they have designed a better agent until they interactively test a module with users. Section 4 discusses user issues.

Curry et al~\shortcite{curry2017review} review assessment techniques for social dialog systems. As mentioned above,  PARADISE~\cite{Walker1997PARADISEAF} uses task completion while optimizing for the number of utterances. This measure has withstood the test of time. It is more relevant for slot-filling systems than for chat and gaming. The measure of response delay does, however, cover all three types of agents. 

\begin{figure}[ht]
   \centering
   \includegraphics[width=0.5\textwidth]{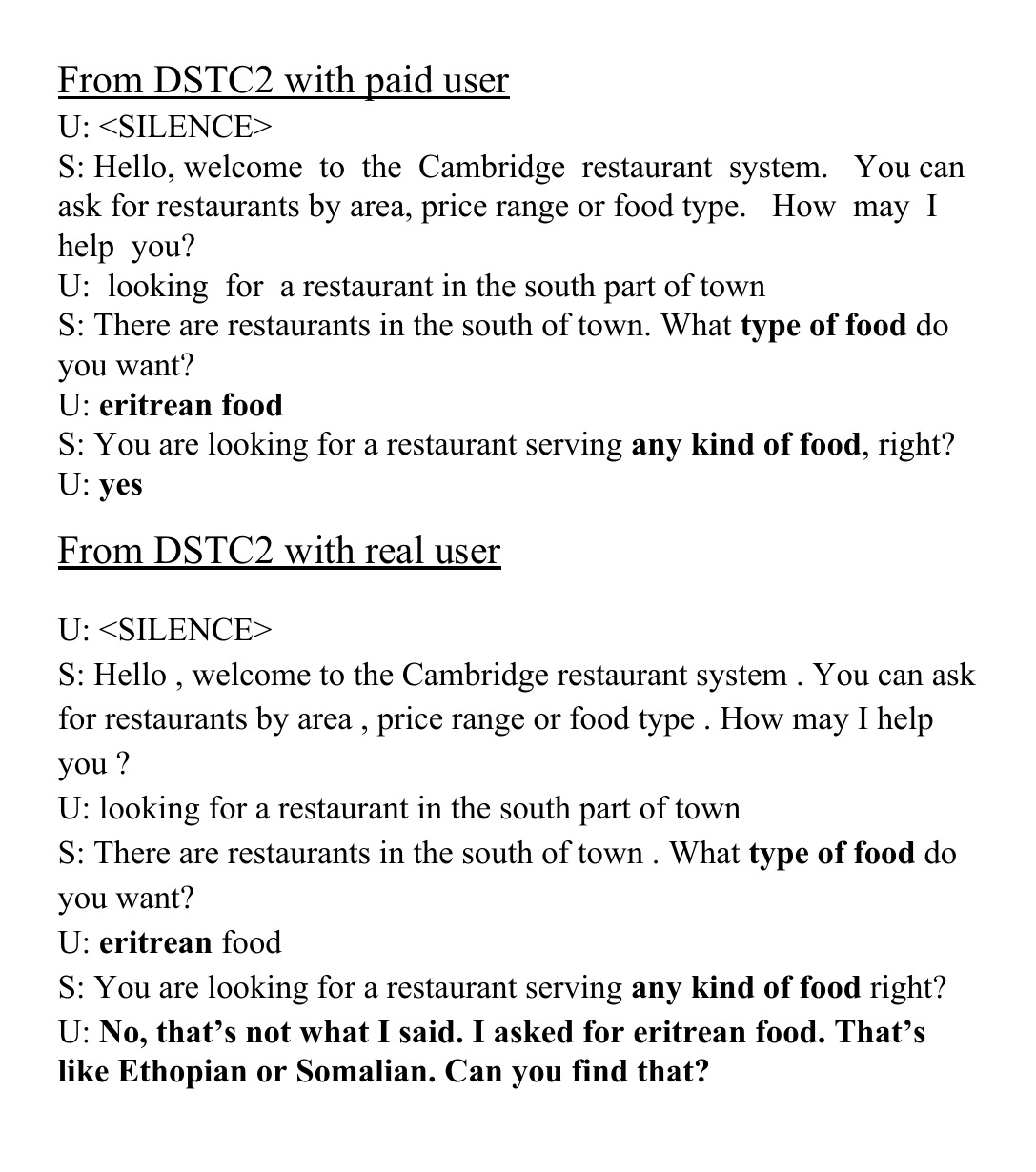}
 \caption{Two example dailogs from DSTC 2 with paid and real users.}
 \label{fig:diff}
\end{figure}

One strategy for agent assessment is to employ third party human judgment~\cite{lowe2017towards}. In this paper, human judges are recruited from Amazon Mechanical Turk (AMT) to determine the validity of the agent'’s response based on criteria such as "topicality, informativeness and whether the context required background information to be understandable”". These three elements measure the amount of relevant material that was provided. This may be helpful to the user. Yet there would be more information about its usefulness if the user had been one of the judges. They also used human-generated responses to determine whether the agent'’s response was useful. The use of real versus paid users, as mentioned in Ai et al~\shortcite{ai2007comparing} could also have made the results more informative. Paid workers are not rewarded for producing a correct outcome no matter how long it takes (Figure \ref{fig:diff}). They are rewarded for simply completing the task. There is also no reward for giving the system additional information. A real user often provides some more context in order to help steer the system to a successful outcome. A paid user will abandon their original goal if it means that they must have a longer dialog with the agent.

Shah et al~\shortcite{shah2016interactive} demonstrated the possibility of interactively adapting to the user’'s response. They used reinforcement learning to enable the system developer to indicate the desired end states. The goal was to endow the system with speech that is more varied than the speech in the training database. By actively modifying the agent using the simulated user turn that follows a given system turn, the new system response became increasingly more appropriate and useful to the user over time. We examine other implementations of this approach in Section 5.

Tan et al~\shortcite{tan2018multimodal} devised a way (using LSTMs) to predict the empathetic response of the user over time, concentrating on the emotions of the user rather than those of the agent. They used the OMG-Empathy dataset, drawing inspiration from the EmotiW Challenge, both of which are oriented toward the \textit{user's} emotions. 

Hancock et al~\shortcite{hancock2019} used a configuration that enabled the user of a deployed agent to give it immediate feedback when it made an error. The agent first had the ability to detect that the user was not satisfied. Then it generated an appropriate question that got the user to explain what it should have said instead. Finally it modelled the user answer to generate future utterances. The paper shows an improvement in performance on an offline test. The user's answers to the agent's request can vary significantly. It is probable that the system may not be able to deal with all of the different types of responses. Yet this is one step toward taking the user into account. We suggest a more direct form of user feedback in Section 5.

Muralidharan et al~\shortcite{muralidharanleveraging} speeds up data annotation of user engagement by capturing specific signals that reflect user engagement. They then use multitask learning and validation to increase labelling accuracy. They choose to create coarse grain labels, which are easier to obtain and are deemed to be more reliable. These labels are then used to generate finer grain labels. This technique was applied to requests for songs.

\subsection{User-oriented guidelines from industry}

The Alexa and Google Assistant ecosystems host many third-party applications that users can converse with, typically as a means of completing a particular task (e.g., ordering food from specific vendors, controlling smart home devices, etc.). In order to ensure a consistent and high-quality user experience, Amazon and Google have each developed a list of development guidelines.

Table \ref{guidelines} lists these guidelines; the first column shows instructions\footnote{\url{https://designguidelines.withgoogle.com/conversation/style-guide/language.html}} that Google provides developers building Google Assistant \textit{actions}  and the second shows instructions\footnote{\url{https://developer.amazon.com/docs/alexa-design/design-voice.html}} that Amazon provides for Alexa \textit{skill} developers. 

\begin{table*}[ht]
\centering
\small
{\renewcommand{\arraystretch}{1.1}
\begin{tabular}{|p{8cm}|p{8cm}|}
\hline
\multicolumn{1}{|c|}{\textbf{Google Assistant}}                                                                                                                                                                                                                                      & \multicolumn{1}{c|}{\textbf{Amazon Alexa}}                                                                                                                                                                                                                                                                      \\ \hline
\multicolumn{2}{|c|}{\textbf{Brevity}}                                                                                                                                                                                                                                                                                                                                                                                                                                                                                                                                                                 \\ \hline
\textbf{Don't launch into monologues. }The system should not dominate the conversation; it should maintain concise responses and allow users to take their turn.                                                                                                                                  & \textbf{Be brief. }The system should minimize the number of steps it takes to complete a task, and avoid redundancy in messages.                                                                                                                                                                                         \\ \hline
\multicolumn{2}{|c|}{\textbf{Simplicity of Word Choice}}                                                                                                                                                                                                                                                                                                                                                                                                                                                                                                                                               \\ \hline
\textbf{Use short, simple words.} Responses should be simple, with plain and basic words, to ensure accessibility to people of all backgrounds.                                                                                                                                               & \textbf{Be informal. }Responses should use a relaxed word choice albeit with a respectful tone.                                                                                                                                                                                                                          \\
\textbf{Avoid jargon and legalese.} Responses should be simple, and the system should avoid specialized expressions that can elicit mistrust or confusion.                                                                                                                                    &                                                                                                                                                                                                                                                                                                                 \\ \hline
\multicolumn{2}{|c|}{\textbf{Natural Dialog}}                                                                                                                                                                                                                                                                                                                                                                                                                                                                                                                                                          \\ \hline
\textbf{Randomize prompts when appropriate. }Responses should be kept diverse and varied, to ensure the conversation remains natural.                                                                                                                                                         & \textbf{Vary responses. }Responses should be randomly selected in order to sound natural and avoid sounding robotic.                                                                                                                                                                                                     \\
\textbf{Avoid niceties. }Formalities should be avoided, and the conversation should remain friendly and informal.                                                                                                                                                                             & \textbf{Use natural prosody. }Alexa skills should mimic the prosody of natural speech to reduce ambiguity, and avoid sounding unnatural or robotic.                                                                                                                                                                      \\
\textbf{Use contractions. }The expansion of contractions should be avoided, as they sound punishing and harsh.                                                                                                                                                                                & \textbf{Use contractions. }Contractions should be used, in order to sound natural and mimic natural conversation.                                                                                                                                                                                                        \\
                                                                                                                                                                                                                                                                                     &\textbf{Engage the user.} The system should prompt the user with simple, open-ended questions. Rather than telling the user exactly what to say, questions should be kept natural.                                                                                                                                       \\ \hline
\multicolumn{2}{|c|}{\textbf{User-Centric Dialog}}                                                                                                                                                                                                                                                                                                                                                                                                                                                                                                                                                     \\ \hline
\textbf{Focus on the user.} Everything should be phrased in the context of the user, rather than making the system persona the focus. Developers should avoid referring to the system (e.g., ‘I placed that order’) and instead focus on the user (e.g., ‘your order will be arriving shortly’). & \textbf{Be engaging.} The system should contain a welcome message which broadly informs the user about its capabilities. It should also use brief question prompts, re-prompt the user in the event of a mistake, offer help in the event of user confusion, have engaging follow-up prompts and a natural exit message. \\
\textbf{Lead with benefits.} When asking the user to perform an action, the response should begin by providing a clear motivation.                                                                                                                                                            & \textbf{Be contextually relevant.} Options should be listed in order of relevance to the ongoing conversation.                                                                                                                                                                                                           \\ \hline
\multicolumn{2}{|c|}{\textbf{Miscellaneous}}                                                                                                                                                                                                                                                                                                                                                                                                                                                                                                                                                           \\ \hline
\textbf{Don't provide UI-specific directions.} The system should avoid providing UI-specific instructions, to better remain relevant in the face of product and interface evolution.                                                                                                          &                                                                                                                                                                                                                                                                                                                 \\ \hline
\end{tabular}}
\caption{A description of user guidelines provided by Amazon and Google for third-party developers for Alexa skills and Google Assistant actions, respectively. They are grouped these into broad categories and provided a clear description for each instruction. The bold guideline titles are taken verbatim from the aforementioned websites. We made an effort to use similar terminology for the instructions to better reflect the motivations.}
\label{guidelines}
\end{table*}

Third-party developers can easily understand these instructions. Although they appear to be formulated in order to maximize user \textit{satisfaction}, upon closer examination the focus is actually on producing \textit{``good'' system output}, thus turned toward the qualities of the \textit{agent}, rather than toward \textit{user satisfaction}.

Multiple guidelines that are common to both sets of instructions, lead to a more ``natural'' and less ''robotic'' appearance. The underlying assumption is that users would prefer assistance in a human-like manner. It is unclear whether these instructions result from real user studies or large scale data analysis. In order to cover every user and each action or skill, a user-centric system should have a flexible set of guidelines that adapts to the nature of each action or skill and to what really satisfies the user. The agent should be able to modify its behavior according to the context. For example, it is not necessarily correct to consistently use a simple vocabulary, particularly if the user is speaking in ``jargon and legalese''. Likewise, if a user appears to be annoyed when the agent repeatedly attempts to be engaging, or ``leading with benefits'', the system should recognize the dissatisfaction and appropriately alter its behavior.

\section{The user issue}
In this Section, we address the elephant in the room: user data. Getting users for training and testing our agents, especially \textit{real} users, is costly in time, in payment and in recruitment effort~\cite{lee2017dialport,lowe2017towards}. Some of the costs include finding the appropriate users, advertising, website management (making the offering attractive), app creation and creating and maintaining hardware and software other than the agent software itself (for example, servers so that the agent is up and running 24/7 and that many users can access the system at the same time). There are also IRB (Institutional Review Board) and privacy concerns. However, real users challenge our research in ways that all others cannot. Real users~\cite{ai2007comparing} do not accept wrong system answers. They do not accept correct answers that do not respond to their needs. Real users will not use a gaming system that has made an error in calculating their score. Real users will not use a chatbot that repeats the same thing several times, even if it was a good answer the first time it was said.

Many researchers have devised alternative data gathering and assessment schemes in an effort to reduce costs. In the end, to truly assess whether an agent is the user's \textit{partner}, we have to bite the bullet and find a stream of real users to interact with our agents.

Based on our experience with Let's Go~\cite{raux2006doing}, we define real users as people who have some personal need for what the agent offers. They find some extrinsic value: getting a task done, enjoyment or companionship. They have some self-defined goal: the reason why they sought to interact with the agent. The real user can be someone who had been a paid crowdsourcing worker but then found that the agent was useful and came back on their own to talk to it again. A real user can also be someone who has come to depend on a given agent to get some task done, like the iPhone for telling time. That is the only agent they use each time they need to do that specific task.

Given the high cost of real users, some less onerous solutions in the literature are: self-dialog, simulated users, crowdsourcing. 
\begin{itemize}
    \item In self-dialog a person imagines a conversation and plays the roles of both interlocutors~\cite{fainberg2018talking}. This obtains data at half the cost and in less time. Much of the nature of human interaction is lost in this approach, for example, the natural entrainment that happens in a dialog between two humans and the exchange of differing ideas. A single-user dialog can never reflect varied word choice or other linguistic characteristics that result from two individuals with different backgrounds interacting with one another. And it cannot measure user satisfaction.
    \item Others, including some of the present authors~\cite{Zhao2018ZeroShotDG} have created simulated users trained on some dataset. This produces a large amount of data at low cost~\cite{moller2006memo}. However, the resulting dialog is limited to exactly what the dataset contained. Generalization to unseen dialog context and novel user expressions is limited or absent. And there cannot be explicit user feedback. 
    \item Many have used crowdsourcing~\cite{eskenazi2013crowdsourcing} to generate human/computer dialog. The Wizard-of-Oz (WoZ) scenario where a human plays the part of the agent~\cite{bonial2017laying} can be used here and requires less system development. While this approach is less onerous (the crowdsourcing platform finds and pays the workers, and the cost is somewhat lower) than attracting real users, it still poses the paid vs real user issue. As mentioned above, the worker is paid to finish a task. Their goal is not altruistic. They want to get many tasks done as quickly as possible in order to earn a decent wage. Thus, they will accomplish the task they are assigned, but they will not insist on a specific outcome even if it is part of their scenario nor will they pursue other options if the system cannot satisfy their request. Figure~\ref{fig:diff} shows a dialog in the DSTC 2/bAbI database test set~\cite{henderson2014second,bordes2016learning}. It illustrates how real users respond in ways that are very different from paid users. We see that the paid user, faced with an incorrect agent response (not what was requested) accepts it. We constructed the second example to illustrate what a real user might have said. 
    
    Crowd workers can become real users if they discover that the agent that they were paid to test is useful. They may then want to return to interact with it again without being paid. We have observed this for DialPort users (see below). The paid task was the opportunity for workers to kick the tires of the Portal, trying various services (remote agents) and finding one that is useful to them. The return visit was counted as a real user dialog. 
\end{itemize}

Industry has real users. Amazon, Google, Microsoft, Apple, etc create very large streams of data from users talking to their agents. However, they cannot share the data (even the Alexa Challenge only shares the data with the individual participating team rather than the public), thus there is no way to replicate results to verify their findings. Furthermore, the data collected (sometimes just the first user utterance after the wake word~\cite{salvador2016wake} or the request to play a song~\cite{muralidharanleveraging} may be very specific. 

One of the first datasets of real user/agent dialogs is the Let’'s Go bus information system~\cite{raux2006doing}. The Let’'s Go dataset contains over 170,000 dialogs collected over 11 years of live daily service to real users via the Port Authority of Allegheny County's phone line. Subsets of Let’'s Go have been used in several challenges, for example, DSTC1 (DSTC1). The complete dataset is available on Github~\footnote{\url{https://github.com/DialRC/LetsGoDataset}}.

\begin{figure}[ht]
   \centering
   \includegraphics[width=0.45\textwidth]{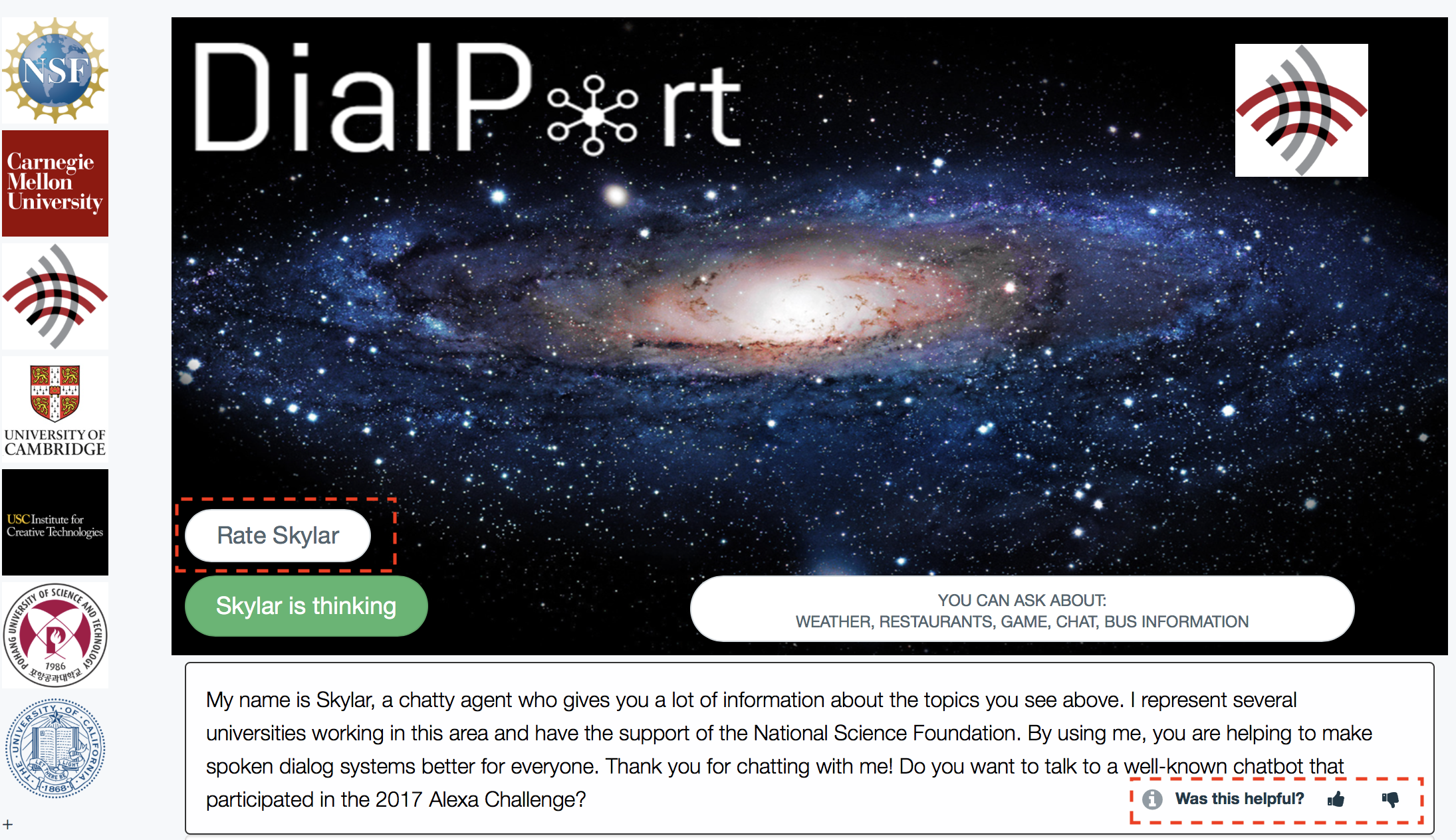}
 \caption{The web interface of DialPort Portal. Users can give both turn-level and dialog-level feedback via thumbs up/down (at lower right).}
 \label{fig:portal}
\end{figure}

DialPort (see the screenshot in Figure~\ref{fig:portal}) collects real user data. Agents from any site anywhere in the world can easily connect to the DialPort Portal using an API call~\cite{lee2017dialport}. Several sites (Cambridge University, USC, UCSC) are already connected and more are preparing to connect. Unlike any other service, connected systems collect real user data in \textit{both text and speech signal format}. The DialPort team advertises the Portal to potential users, maintains it and tallies its throughput. A daily report displays data from testers and real users, not developers, (second, third, etc usage, determined from the IP address). DialPort is a hybrid system including both slot-filling systems (Cambridge) and chatbots (UCSC's Slugbot and CMU’s Qubot). In DialPort, Qubot deals with all \textit{out of domain} (OOD) user utterances. Qubot keeps the dialog moving along for turns where none of the Portal's agents is capable of responding. This keeps the system from being unresponsive to the user. Another way that DialPort keeps the dialog moving along is its use of \textit{backup systems}. In the rare case when Cambridge, for example, is not available, the CMU in-house Let's Eat restaurant agent answers the user. Thus, the dialog has continuity. The user never hears ''The Cambridge system is not responding at present, can I help you with something else?''. The CMU Let's Guess movie game is the backup for USC’s Mr. Clue game. UCSC’s Slugbot has Qubot as its backup. A non-CMU system is always chosen by the Portal over the CMU backup system when that system is available.

If researchers want to start a first phase of their assessment with crowdsourced workers, the CMU Dialog Research Center (DialRC) offers DialCrowd, a layer between the developer and the crowdsourcing platform of choice~\citep{lee2017dialport}. This layer helps developers create effective instructions, set up commonly used testing strategies (like ABX tests~\citep{munson1950standardizing}), determine correct payment, control incoming data quality and get a first set of statistics.

With logfile analysis, we can determine if the user returns to use the agent again, how frequently this happens and the length of their engagement. In the case of CMU's Qubot, we assess its contribution by determining whether Qubot'’s turn allowed the conversation to continue seamlessly as opposed to the user quitting or getting frustrated.

\section{Toward user-based assessment}
This section proposes ideas for user-centric assessment. This first step is intended to inspire the community, who can extend and elaborate what we propose according to their specific goals. 

\subsection{Thumbs up}
One of the easiest ways to determine whether an agent's output is of value is to display thumbs up/thumbs down icons on a webpage and ask for the user's opinion at the level of both the individual turns and the overall dialog~\cite{young2017}. Figure~\ref{fig:portal} shows the these icons in the lower right corner on DialPort's website. The user's opinion is gathered immediately as the interaction takes place, rather than relying on the user's memory once the dialog is finished.

\subsection{Flow}
\label{sec:flow}
Many user-centric metrics focus on whether the user found a conversation to be engaging \citep{wozuseroriented} or whether they would want to use the system again (e.g., the Amazon Alexa Prize Challenge, \cite{alexaprize}). These metrics are turned toward the user. However, some are applicable to non-task-oriented systems only.
Task oriented systems are typically assessed automatically using entity accuracy and through evaluation of the output language with metrics such as perplexity and BLEU. These metrics mostly demonstrate the prowess of the developer in creating the agent and do not take the user into account. 

A more user-centric metric is \textit{dialog flow}. We posit that this metric reveals the coherent nature of communication with a given agent. It is applicable to both task-oriented and non-task-oriented systems. After defining dialog flow, we will discuss how the term has been used before, the relation of \textit{dialog flow} to other previously-used user-centric metrics and its value as shown in the example of senior users.  
 
We define \textit{dialog flow} to be a measure of whether a conversation is ``going smoothly.'' At each step of the conversation, this metric measures whether the system response is a logical continuation of the dialog history.

 
This term has been used previously in the literature, however there is a lack of agreement about its meaning.
For some authors, dialog flow reflects whether the system has stayed on topic and how topics have changed over the course of the dialog \citep{flowtopics}. However, this definition only takes into account the continuity of the topics discussed and not whether the dialog was smooth, for example, if the system repeated the same utterance three times, it is considered to be one topic but each successive repetition of that turn is uninformative. Other authors examined whether the agent could follow a set of predefined scripts, which were referred to as the dialog flow \citep{dialogflowscripts1, dialogflowscripts2}. While this approach describes the dialog, it does not give us a way to assess it. The assumption that there is a predefined script, may be valid rigidly hand-crafted task-oriented systems but not for non-task-oriented systems.

Our definition of dialog flow complements other user-centric metrics. The Alexa Prize Challenge \citep{alexaprize} acknowledges the user in its assessment metrics. All of the contestants were evaluated based on real user feedback. After conversing with a particular system, the user was asked to rate whether they would want to talk to the system again on a scale of 1 to 5 \citep{alexaprize}. A subset of the users was asked to rate how engaging the system was. Other characteristics of the systems such as coherence and conversational depth were measured automatically \citep{alexaprize}. Coherence and depth do contribute to an impression of dialog flow. This can complement data on whether the user would return for another conversation. To further measure flow, the user could be asked (similarly to thumbs up) how content they were at each turn of the conversation, rather than simply after the entire dialog.

In the context of task-oriented dialogs, flow can reflect whether the user is closer to completing the task after each turn. In the context of non-task-oriented dialogs, flow can capture whether the dialog is coherent and smooth from the user's point of view. Both completing a task and dialog smoothness may be captured in the following user turn. 

Dialog flow advances the role of the user as both an active and passive partner and provides a global assessment of the partnership. This more versatile approach to partnership would be beneficial, for example, for seniors since it intrinsically considers the features described in section \ref{subsec:seniors}. If a dialog flows well, the dialog will proceed in a step-by-step fashion which gives the user a feeling of more control over the course of the dialog.

\subsection{Next-turn User Feedback (NUF)}
In this section we examine the effect of looking at the user utterance that \textit{follows} the system response. We propose a method to effectively leverage this additional information. One of the standard assessment methods used in the literature is turn-level system response evaluation where, for each dialog context, we determine whether the model's system response output matches the ground truth system response. Metrics have been proposed to test various aspects of performance, like lexical similarity (BLEU), embedding-based semantic matching etc. 

Prior research has shown that for open-domain chatting these metrics correlate poorly with human evaluation~\cite{liu2016not}. Moreover, prior research has demonstrated that the context-to-response problem is one-to-many in an open domain environment. Thus, considering each response in a dataset to be ground truth gives us noisy supervision signals, which is one of the causes of the dull-response problem~\cite{zhao2017learning}, i.e. the model can only learn to generate dull, generic responses. 

The NeurIPs2018 ConvAI challenge \citep{convai2} further highlighted the discrepancy between automated metrics and human evaluation. \textit{Hugging Face} topped the leaderboard in every automated metric (PPL, hits@1, F-1), however the final measure in the competition used human evaluation \citep{emily2018}. The winner was \textit{Lost in Conversation}, a team that did not do particularly well on the automated metrics and just barely made it into the finals. The results of this challenge stress the importance of constructing well-defined automated metrics, that better correlate with a human evaluation of conversation quality.

\begin{table}[ht]
    \centering
    \begin{tabular}{p{0.45\textwidth}} \hline 
    \textbf{Context} \\
    usr: What's the Mountaineering Club doing this weekend? \\
    sys: We are going to climb a cliff on the Colorado River. \\
    usr:  How hard is the climb?  \\ \hline
    \textbf{System Response} $S_{sys}=5$\\ 
    It is a very difficult climb because there are not many places to put your hands and feet. \\ \hline
    \textbf{Next-turn User Response} $S_{usr}=3$ \\
    What did you do last week ?\\\hline
    \end{tabular}
    \caption{Example where the the system response is appropriate given the context, but the user response suggests that the other speaker is not satisfied. This example comes from Daily Dialog corpus.}
    \label{tbl:example-1}
\end{table}

\begin{table}[ht]
    \centering
    \begin{tabular}{p{0.45\textwidth}} \hline 
    \textbf{Context} \\
    sys: Welcome to the CMU Let's Go bus information system. To get help at any time, just say Help or press zero. What can I do for you? \\
    usr: next bus from 6c u 12a\\
    sys: I think you want the next bus. Am I right?\\
    usr: yes\\
    sys: Alright Where do you wanna leave from?\\
    usr: eight\\
    \textbf{System Response} $S_{sys}=2$\\ 
    For example. You can say, when the next bus from CMU to the airport? \\ \hline
    \textbf{Next-turn User Response} $S_{usr}=5$ \\
    bigelow boulevard\\\hline
    \end{tabular}
    \caption{Example where the the system response seems off, but the user is in fact satisfied. This example comes from DSTC-1.}
    \label{tbl:example-2}
\end{table}

We would like to develop a measure of whether the agent utterance was correct as defined by whether the user would approve of it. We concentrate on the validity of the agent's utterance conditioned on the \textit{following} user utterance, rather than the agent's utterance given the \textit{preceding} utterances alone. Although ideally it is true that the entire dialog that follows a system response should be used to judge the response, this becomes prohibitively expensive when used to assess each system utterance. Therefore, we assume that the next-turn user reply/feedback (which we call NUF) provides sufficient supervision signals for our goal. Given this assumption, prior work has used coherence-based measures to determine whether a system response is on topic~\cite{li2016deep}. We cannot simply predict the user's response since we would be predicting both good and bad (``that’s right'' and ``No, you didn't understand!'') with no reward for predicting the user response that reflects success. Here is an example:
\begin{itemize}
    \item \textbf{Sys:} ``ok, leaving from Forbes and Shady, where are you going?''
    \item \textbf{Usr:} ``Not Forbes and Shady! Forbes and Murray''
\end{itemize}

In this example, the two utterances have significant word overlap. Thus, automatic analysis would find that the agent utterance is on topic and correct. But it is evident that the agent utterance is incorrect. It does not respond to the user's request and will not result in useful information. Therefore, we propose NUF, a new extension of standard system response evaluation that takes the following user response into account during assessment. Specifically, standard system response evaluation asks ``is response $x$ an appropriate response given the context $c$?''. Instead, our metric asks ``has response $x$ \textit{satisfied} the user based on context $c$ and the following user reply $u$?''. We define being satisfied in a task-oriented slot-filling domain as simply whether the system provides useful information to the user. In more complex chat-based systems, the definition of flow in Section~\ref{sec:flow} can be applied. 

\subsubsection{Human Study of NUF}
In order to test our hypothesis, a human study compares the ability of human annotators to assess the quality of a system utterance in two different situations: first given just the context and the system utterance alone and second given the context, the system utterance \textit{plus} the following user response. We used four datasets in order to cover both slot-filling and non-slot-filling systems: DSTC-1~\cite{williams2013dialog}, DSTC-2~\cite{henderson2014second}, Daily Dialog~\cite{li2017dailydialog} and Switchboard~\cite{godfrey1992switchboard}. 30 \textit{c-x-u} tuples from each dataset were sampled and labelled by four expert raters (the coauthors of this paper) who gave a score of 1 to 5, ranging from no user satisfaction to complete user satisfaction. Appendix~\ref{sec:appendix} contains the annotation instructions. 
\begin{table}[ht]
    \centering
    \begin{tabular}{p{0.16\textwidth}|p{0.12\textwidth}p{0.12\textwidth}} \hline 
    Kappa          & $S_{sys}$ & $S_{usr}$ \\ \hline
    DSTC-1         & 0.359     & 0.616 \\
    DSTC-2         & 0.487     & 0.593 \\ 
    Daily Dialog   & 0.196     & 0.287 \\
    Switchboard    & 0.172     & 0.348 \\ \hline
    \end{tabular}
    \caption{Fleiss Kappa among the four raters.}
    \label{tbl:kappa}
\end{table}
Table~\ref{tbl:kappa} shows that the rater agreement on $S_{usr}$ (taking the following user utterance into account) is much higher than that for $S_{sys}$ for all four datasets. This confirms the hypothesis that, even for humans, it is difficult to judge whether a response is truly appropriate if \textit{only} the previous dialog history (dialog context) is used. But, given the following user turn, it becomes significantly easier. This result also implies that the following user turn contains salient information that helps distinguish between ``good'' system responses versus ``mediocre'' or ``bad'' responses, which are rarely used in standard supervised-learning-based training for end-to-end neural dialog systems. 

Table~\ref{tbl:example-1} and Table~\ref{tbl:example-2} show examples of dialogs where $S_{sys} > S_{usr}$ and vice versa. They show that the appropriateness of a system response is much clearer when we see the following user utterance. For example, the dialog in Table~\ref{tbl:example-1} shows that although the system gives an on-point explanation for the previous question, the user is apparently not interested in continuing down that path of the conversation. 

\begin{table}[ht]
    \centering
    \begin{tabular}{p{0.12\textwidth}|p{0.12\textwidth}|p{0.12\textwidth}} \hline 
     $S_{sys} = S_{usr}$ & $S_{sys} < S_{usr}$ & $S_{sys} > S_{usr}$ \\ \hline
    46.3\% & 18.6\% & 35.0\% \\ \hline
    \end{tabular}
    \caption{Percentage of the data where the annotators underestimated and overestimated the quality of system response relative to what was indicated by the \textit{following} user utterance.}
    \label{tbl:compare}
\end{table}

Table~\ref{tbl:compare} compares the annotations of conversation quality both in the presence and in the absence of the \textit{following user utterance.} We show the percentage of the data where the raters gave the system higher scores than the user did and inversely where the raters' score was lower than that of the users. This is equivalent to comparing the annotators' perception of the system utterance to the user's perception of the system utterance. We find that the difference (practically double) between $S_{sys} < S_{usr}$  and $S_{sys} > S_{usr}$ underlines our claim that judging system \textit{correctness} should be based on the following user utterance. Using the system utterance plus previous context alone leads to an artificially high assessment of the quality of the system utterance and consequently of system performance.

\subsubsection{Automatic NUF Prediction of $S_{usr}$}
We wanted to see how we can automatically predict $S_{usr}$ given \textit{c-x-u}. 

\textbf{Data Annotation}
The four experts annotated a subset of the DSTC-1 dataset, resulting in a total of 1250 \textit{c-x-u} data points with 150 overlapped (seen by all annotators) to ensure inter-rater agreement. The Fleiss kappa was computed, giving 0.207 for $S_{sys}$ and 0.515 for $S_{usr}$. Then both classification and regression models were created. 70\% of the data was used for training and the remaining 30\% was used to test. 

\textbf{Model Details}
The last turn in the dialog context history, the system response and the user response are turned into vector representations via bag-of-ngram features for $n \in [1, 2]$ with TF-IDF (term frequency-inverse document frequency) weighting. We combined these sources of information by simply concatenating the feature vectors. Also, for classification, a support-vector-machine (SVM) with a linear kernel is used, as well as Ridge regression (linear regression with $L_2$ regularization). Given the above setup, we are interested in how well can we predict $S_{usr}$ and what the input features are. 

\begin{figure}[ht]
   \centering
   \includegraphics[width=0.45\textwidth]{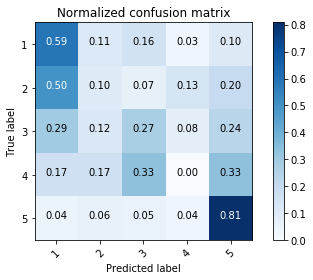}
 \caption{Confusion matrix for classification model with c-x-u as inputs.}
 \label{fig:confuse}
\end{figure}

\textbf{Results}
\begin{table}[ht]
    \centering
    \begin{tabular}{p{0.16\textwidth}|p{0.12\textwidth}p{0.12\textwidth}} \hline 
    Input       & Acc       & MAE\\ \hline
    $c$         & 47.5\%     & 1.31 \\ 
    $x$         & 55.7\%     & 1.14 \\ 
    $u$         & 60.3\%     & 0.81 \\ \hline
    $c,x$       & 54.6\%     & 1.08 \\ 
    $c,u$       & 62.9\%     & 0.78 \\ 
    $x,u$       & 65.8\%     & 0.67 \\ \hline
    $c,x,u$     & 65.8\%     & 0.68 \\ \hline
    \end{tabular}
    \caption{Results for $S_{usr}$ prediction. MAE stands for absolute mean error.}
    \label{tbl:prediction}
\end{table}
Table~\ref{tbl:prediction} shows the results. The best performing model is able to achieve 65.8\% accuracy and MAE 0.67, which represents less than 1 error to the ground truth expectation. Further, the results yield the following insights:
\begin{itemize}
    \item The most useful features to predict $S_{usr}$ stress the importance of understanding user feedback. Table~\ref{tbl:prediction} shows that the model struggles when only presented with the context and the system response whereas it achieves strong performance even when it only uses $u$ as its input. This opens up the possibility of having a universal user satisfaction model that solely takes the user response as its input. This is easier to obtain for training than manually labelling every context-response-user response tuple. 
    \item Regression is more suitable than classification for this task. Figure~\ref{fig:confuse} shows that the classification models mostly confuse close-by classes, e.g. class-1 and class-2, which come from the raters' ambiguity concerning similar classes. Therefore, regression naturally takes care of the ordinal nature of $S_{usr}$. 
\end{itemize}

\section{Conclusions and future directions}
This paper is a first step toward making the agent into a real partner for human users. It suggests several guidelines for system developers: avoid paid users in favor of real ones, use an interface like DialPort to collect data, do active rather than off-line assessment, ask for immediate feedback (e.g. thumbs up), care about the flow of the dialog, use NUF or create measures similar to it. It shows how we can quantify the effect of the agent on the user and serves as a point of departure for the development of more assessment strategies that turn toward the user. 

\bibliography{acl2019}
\bibliographystyle{acl_natbib}

\appendix
\section{Appendix: Supplemental Material}
\label{sec:appendix}
\textbf{Annotation instructions for $S_{sys}$}
Score (sys) given the dialog context and the system response ONLY, use a five point Likert scale to judge if the system response gives useful information and matches the flow of the conversation. You are the observer, judging if YOU think, in YOUR OPINION, the system output was 1, 2. 3, 4, or 5.
\begin{enumerate}
    \item System response was irrelevant or incorrect.
    \item system response is slightly off topic or giving relevant but inaccurate information.
    \item System response is on topic but neutral, you cannot judge if it's correct or incorrect.
    \item System response is somewhat useful.
    \item System response gives the user exactly what they needed
\end{enumerate}

\textbf{Annotation instruction for $S_{usr}$}
Score (sys+usr) given the dialog context, the system response AND the user response, using a five point Likert scale to judge how  much this system response satisfied the user by giving them information that is useful and correct. You are the observer, reporting what you thought the USER's opinion of the system output was, based on the user's turn.
\begin{enumerate}
    \item System response was judged by the user to be totally incorrect;
    \item  System response was judged by the user to not be what they wanted, but not totally off
    \item The user was neutral about the value of the system response - or given the content of the user utterance, you could not judge the value to the user
    \item System response was judged by the user to be somewhat helpful
    \item System response was judged by the user to be exactly what they needed
\end{enumerate}

\end{document}